\documentclass{article}
\usepackage[T1]{fontenc}
\usepackage[utf8]{inputenc}
\usepackage[]{ismir} % Remove the "submission" option for camera-ready version
\usepackage{amsmath,cite,url}
\usepackage{graphicx}
\usepackage{color}
\usepackage[table,xcdraw]{xcolor}

\usepackage{booktabs}
\usepackage{multirow}
\usepackage{listings}
\usepackage{caption}

\title{MUST-RAG: MUSical Text Question Answering\\with Retrieval Augmented Generation}

\begin{document}

\author{
Daeyong Kwon\textsuperscript{1}\footnotemark[1], 
SeungHeon Doh\textsuperscript{1}\footnotemark[2], 
Juhan Nam\textsuperscript{1}\footnotemark[3]\\
\textsuperscript{1}Graduate School of Culture Technology, KAIST, South Korea
}

\maketitle

\footnotetext[1]{\texttt{ejmj63@kaist.ac.kr}}
\footnotetext[2]{\texttt{seungheondoh@kaist.ac.kr}}
\footnotetext[3]{\texttt{juhan.nam@kaist.ac.kr}}

\begin{abstract}
% Recent advancements in Large language models (LLMs) have demonstrated remarkable capabilities across diverse domains. While they exhibit strong zero-shot performance on various tasks, LLMs' effectiveness in music-related applications remains limited due to the relatively small proportion of music-specific knowledge in their training data. To address this limitation, we propose \textbf{MusT-RAG}, a comprehensive framework based on Retrieval Augmented Generation (RAG) to adapt general-purpose LLMs for text-only music question answering (MQA) tasks. 
% We evaluated RAG performance using MusWikiDB, a music-specialized indexed database designed to support retrieval for MQA.
% Experimental results on both in-domain and out-of-domain MQA benchmarks demonstrate that our proposed approach consistently outperforms existing methods across all settings.
Recent advancements in Large language models (LLMs) have demonstrated remarkable capabilities across diverse domains. While they exhibit strong zero-shot performance on various tasks, LLMs' effectiveness in music-related applications remains limited due to the relatively small proportion of music-specific knowledge in their training data. To address this limitation, we propose \textbf{MusT-RAG}, a comprehensive framework based on Retrieval Augmented Generation (RAG) to adapt general-purpose LLMs for text-only music question answering (MQA) tasks. RAG is a technique that provides external knowledge to LLMs by retrieving relevant context information when generating answers to questions.  To optimize RAG for the music domain, we (1) propose MusWikiDB, a music-specialized vector database for the retrieval stage, and (2) utilizes context information during both inference and fine-tuning processes to effectively transform general-purpose LLMs into music-specific models. Our experiment demonstrates that MusT-RAG significantly outperforms traditional fine-tuning approaches in enhancing LLMs' music domain adaptation capabilities, showing consistent improvements across both in-domain and out-of-domain MQA benchmarks. Additionally, our MusWikiDB proves substantially more effective than general Wikipedia corpora, delivering superior performance and computational efficiency.
% \footnote{The model, dataset, and source code will be open-sourced.}

% As part of this framework, we introduce \textbf{MusWikiDB}, a music-specialized indexed database that enables effective retrieval of relevant passages.
% Experimental results on both in-domain and out-of-domain MQA benchmarks demonstrate that our approach consistently outperforms existing methods across all settings.

%For the music-specific RAG, we introduce \textbf{MusWikiDB}, a specialized vector database containing curated music information, and systematically evaluate design choices for optimal retrieval performance by comparing various embedding models and chunking strategies. 

% The MusT-RAG framework operates in three stages—indexing, retrieval, and generation—leveraging \textbf{MusWikiDB}, a specialized vector database containing curated music information. Our proposed RAG methodology demonstrates greater efficiency for in-domain adaptation compared to traditional fine-tuning approaches, while also showing superior capability in handling new knowledge in out-of-domain scenarios.\footnote{The model, dataset, and source code will be open-sourced.}
 % across various text-only MQA benchmarks.

\end{abstract}

\section{Introduction}\label{sec:introduction}

\begin{figure*}[h]
  \centering
  \includegraphics[width=0.9\linewidth]{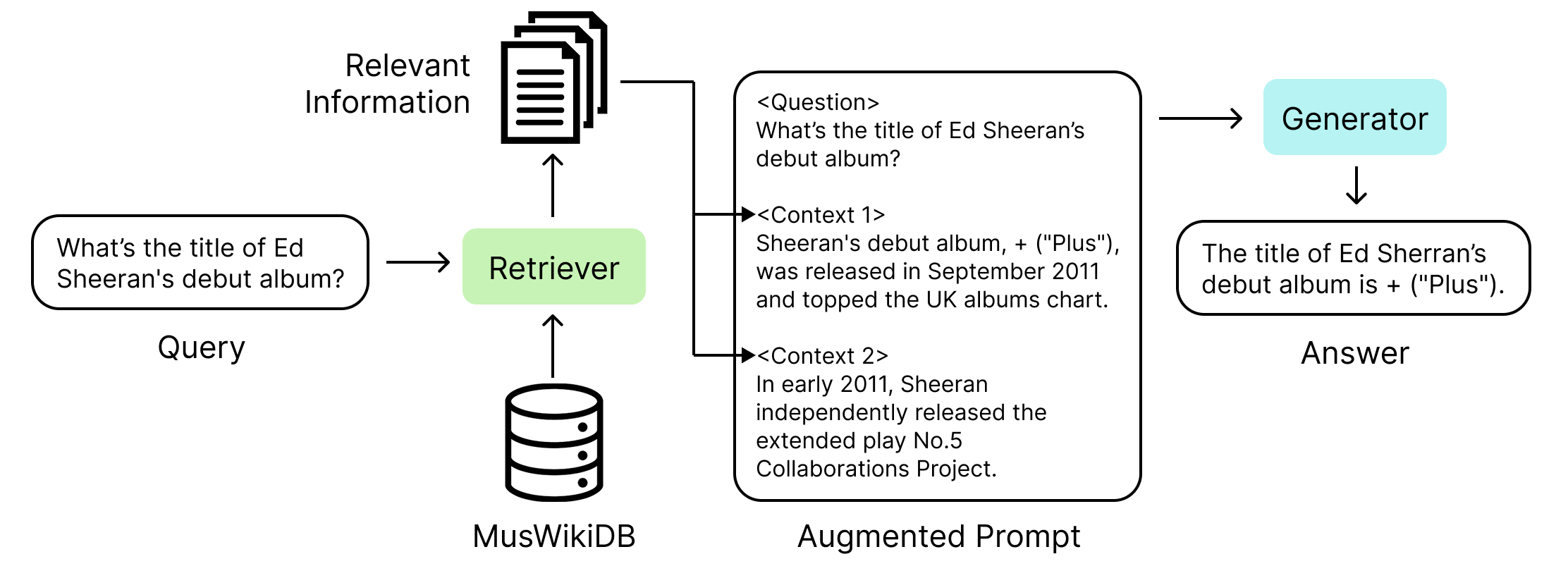}
  \caption{Overview of our \textbf{MusT-RAG} framework. The retriever searches for relevant information in \textbf{MusWikiDB} based on similarity for music-related queries, and augments the generator's prompt with this information to generate an answer.}
  \label{fig:main_figure}
  \vspace{-3mm}
\end{figure*}

% Text-only Music QA의 필요성
Recent advancements in Large language models (LLMs) have demonstrated impressive capabilities across a wide range of tasks, thanks to their massive scale and ability to generalize across diverse domains. 
However, LLMs still face significant limitations in music-related applications due to the relatively small amount of music-specific knowledge in their training data. To effectively deploy general LLMs in music-related domains such as music recommendation systems and chatbots, a deep understanding of Music Question Answering (MQA) in text-only settings is essential. Mastering text-based MQA would enable LLMs to provide more accurate and contextually aware responses to user questions about music, ultimately enhancing the user experience in music-related applications. Developing a robust text-only music QA framework is therefore a key step toward improving the adaptability of LLMs in the music domain.

% 기존 도메인 adaptation을 위한 fine-tuning 방법의 한계
Traditionally, domain adaptation of LLMs has often been achieved by fine-tuning them on domain-specific data~\cite{jeong2024fine, sahoo2024large, satterfield2024fine}. However, this approach faces challenges in securing high-quality training data, and as model size increases, the training time and cost also rise significantly. Additionally, continuously updating the model with new knowledge remains a persistent challenge.

%Recent advancements in Large Language Models (LLMs) have shown impressive capabilities in a wide range of tasks due to their massive scale and ability to generalize across diverse domains.
%Fine-tuning these models on domain-specific data has led to notable performance improvements, particularly in specialized areas such as medical and legal domains~\cite{jeong2024fine, sahoo2024large, satterfield2024fine}.
%However, when it comes to music-related applications, LLMs still face limitations due to the relatively small amount of music-specific knowledge in their training data.
%Moreover, fine-tuning, while effective, comes with significant challenges. As models continue to grow in size, the cost and time associated with fine-tuning increase substantially. Additionally, keeping these models updated with new, domain-specific knowledge presents a continual challenge, as it requires frequent retraining and incurs further computational overhead. 
%Furthermore, for general LLMs to be effectively deployed in music-related domains such as music recommendation systems and chatbots, a deep understanding of music question answering (MQA) in text-only settings is crucial. Mastering text-based MQA would allow LLMs to provide more accurate and contextually aware responses to user queries about music, ultimately improving user experience in music-related applications. Therefore, developing a robust text-only music QA framework represents a key step toward enhancing the adaptability of LLMs in the music domain.

In this paper, we propose \textbf{MusT-RAG}, a framework that leverages Retrieval Augmented Generation (RAG)~\cite{lewis2021rag} techniques to enhance general-purpose LLMs for music-specific tasks. The core idea behind MusT-RAG is to augment LLMs with external knowledge retrieval mechanisms. Specifically, the model retrieves relevant external knowledge from a pre-constructed, comprehensive music-specific vector database in order to answer input questions.

For music-domain specific retrieval, we introduce \textbf{MusWikiDB}, which, to our knowledge, is the first comprehensively curated vector database designed specifically for music-related content. We explore various design choices for optimizing retrieval performance, including embedding models and chunking strategies. By incorporating this retrieval process, MusT-RAG enables LLMs to efficiently generate contextually relevant responses, drawing on specialized music knowledge to enhance performance on music-related tasks, all without requiring additional training. Furthermore, we extend the application of RAG beyond inference by incorporating contextual information during the fine-tuning process. Our empirical analysis reveals that this context-aware training enhances the model's contextual understanding capabilities, defined as the ability to generate coherent and relevant text within a specific context~\cite{zhao2024enhancing}, outperforming conventional fine-tuning approaches.

% Additionally, we identified that the conventional fine-tuning approach significantly degrades contextual understanding—the model's ability to interpret and generate text that is coherent within a given context~\cite{zhao2024enhancing}. To address this, we applied fine-tuning using RAG-style data structured in the format of \textit{(context, question, answer)}, rather than direct QA pairs.

MusT-RAG demonstrated the effectiveness of RAG across all scenarios including in-domain and out-of-domain settings, as well as both fine-tuning and inference stages. By retrieving relevant \textbf{context} information from the database, MusT-RAG effectively addresses the music domain adaptation problem. Our contributions are as follows: \textit{i)} We propose the \textbf{MusT-RAG} framework, which leverages RAG to retrieve relevant context from a music-specific database for answer generation. \textit{ii)} We create \textbf{MusWikiDB}, the first comprehensive music-specific vector database for RAG. \textit{iii)} We demonstrate that RAG-style fine-tuning can resolve the issue of decreased contextual understanding performance with conventional fine-tuning. \textit{iv)} We introduce \textbf{ArtistMus}, a benchmark designed to evaluate artist-related questions in text-only MQA tasks, addressing a gap in existing evaluations.

\section{Music Question Answering}

Question Answering (QA) refers to the task of providing an appropriate answer to a given question, which is one of the Information Retrieval (IR) tasks~\cite{allam2012question}. The task is typically framed as retrieving relevant information from a collection of documents or knowledge sources to answer fact-based questions. Open-domain QA involves answering questions from a vast and varied set of topics using a large collection of general knowledge documents~\cite{rajpurkar2016squad, karpukhin2020dense}. In contrast, domain-specific QA targets specialized fields such as medicine~\cite{pal2022medmcqa}, law~\cite{chalkidis2021lexglue}, or music~\cite{weck2024muchomusic, yuan2024chatmusician, ramoneda2024role, li2024music}, where both the document set and the questions are confined to that domain.
In this work, we define the MQA task as the problem of providing accurate and relevant answers to music-related questions by leveraging domain-specific musical knowledge.

Several recent studies have introduced music-related benchmarks to evaluate LLM performance. MuChoMusic~\cite{weck2024muchomusic} features 1,187 audio-based multiple-choice questions that assess both musical knowledge and reasoning capabilities. MusicTheoryBench~\cite{yuan2024chatmusician} contains 372 expert-validated questions designed to evaluate advanced music knowledge and reasoning skills. TrustMus~\cite{ramoneda2024role} comprises 400 questions across four domains—\textit{People}, \textit{Instruments and Technology}, \textit{Genres, Forms, and Theory}, and \textit{Culture and History}—all derived from The Grove Dictionary Online~\cite{sadie2001new}. ZIQI-Eval~\cite{li2024music} presents a comprehensive evaluation framework consisting of 14,000 comprehension tasks that span 10 major topics and 56 subtopics, encompassing a broad spectrum of music-related knowledge.

% Evaluation of the MQA task relies on music-related benchmarks; however, existing benchmarks are limited in scope. Many focus on audio-based understanding~\cite{weck2024muchomusic} or symbolic music representations such as ABC notation~\cite{li2024music, yuan2024chatmusician}, which are primarily used for music generation or structural analysis. Even among recent text-only MQA benchmarks~\cite{ramoneda2024role, yuan2024chatmusician, li2024music}, the emphasis has been on musicological content—such as harmony, melody, or historical context—rather than rich metadata that captures artist-specific knowledge.

A significant shortcoming of existing benchmarks is their inadequate representation of rich metadata about tracks, artists, and albums—information crucial for everyday music listening contexts. Current text-only QA benchmarks inadequately address common music information needs, lacking comprehensive coverage of details that listeners frequently seek: complete discographies, artist collaboration networks, creative evolution across albums, and notable career achievements. This pronounced disparity between current MQA capabilities and the practical information demands of music consumers highlights the pressing need for benchmarks specifically designed to evaluate responses to artist-centric quetions.

\section{Retrieval Augmented Generation}

\subsection{RAG Framework}

Retrieval-Augmented Generation (RAG) enhances the capabilities of LLMs by combining their generative abilities with access to external knowledge. Instead of relying solely on parametric memory, RAG retrieves relevant passages from an external database during inference time to ground responses in factual context.

\subsubsection{Indexing}

The first step in RAG is constructing a searchable knowledge database. This involves segmenting a large corpus into fixed-size text passages (chunking), followed by representing each passage using an embedding models. Various embedding models can be used for indexing:

\vspace{1mm} \noindent \textbf{Sparse Embeddings}~\cite{robertson1994some,behnamghader2024llm2vec} use term frequency-based scoring to match exact keywords, offering fast and interpretable retrieval for large-scale datasets.

\vspace{1mm} \noindent \textbf{Dense Embeddings}~\cite{devlin2019bert, izacard2021unsupervised, behnamghader2024llm2vec} map questions and documents into a shared vector space, enabling semantic matching beyond keyword overlap.

\vspace{1mm} \noindent \textbf{Audio-Text Joint Embeddings}~\cite{wu2023large, manco2022contrastive, doh2023toward, doh2024enriching, wu2025clamp} extend this further by jointly embedding text with the audio modality. By leveraging contrastive learning between audio and text, they can serve as more domain-specialized text embedding models for music-related tasks.

\subsubsection{Retrieval}

Formally, the retriever \(R\) is defined as a function:
\[
R : (q, D) \rightarrow c
\]
where \( q \) is the input question, \( D \) is the entire database of text passages, and \( c \subset D \) is the filtered context consisting of the top-\( k \) passages, such that \( |c| = k \ll |D| \).
Each passage~\footnote{A passage refers to a portion of a document relevant to a query~\cite{wade2005passage}.} \( p \in D \) is scored based on its similarity to the input question using cosine similarity between their embeddings:
\[
\text{sim}(q, p) = \frac{E(q) \cdot E(p)}{\lVert E(q) \rVert \lVert E(p) \rVert}
\]
Here, \( E(\cdot) \) denotes an embedding function that maps both questions and passages into a shared vector space. The retriever ranks all passages in \( D \) by their similarity scores and selects the top-\( k \) passages to form \( c \), which serve as the external context for the generation step.

\subsubsection{Generation}

The retrieved context \( c \) is provided to a generator LLM, which produces an output sequence using next-token prediction. Each token \( x_i \) is generated conditioned on the input query \( q \), the retrieved context \( c \), and the previously generated tokens \( x_{<i} \):
\[
p(x_1, \dots, x_n \mid q, c) = \prod_{i=1}^{n} p_\theta \left( x_i \mid [q, c; x_{<i}] \right)
\]
This structure enables the model to dynamically incorporate external knowledge during inference, improving factual accuracy and adaptability without retraining.

\subsection{RAG vs. Fine-tuning}

LLMs often struggle with specialized tasks such as MQA due to limited exposure to domain-specific knowledge during pretraining. To address this, two primary domain adaptation strategies are commonly used: fine-tuning and RAG. Fine-tuning is akin to a closed-book exam: the model internalizes domain knowledge during training and must rely solely on that knowledge at inference. While effective for learning structured formats or stylistic patterns~\cite{shuster2021retrieval, borgeaud2022improving}, it is resource-intensive and inflexible when adapting to new or frequently changing knowledge. In contrast, RAG is like an open-book exam: the model dynamically retrieves relevant information from an external knowledge source during inference. This enables LLMs to access up-to-date and specialized information without retraining. Prior studies~\cite{yasunaga2022deep, wang2023self} show that RAG improves factual accuracy, mitigates hallucinations, and provides greater transparency by allowing source verification. It is also more scalable and economically efficient, as it does not require updating model parameters~\cite{borgeaud2022improving}. These benefits are especially useful in rapidly evolving domains like music, where new artists, compositions, and styles continuously emerge.

\subsection{RAG with Fine-tuning}

While fine-tuning typically relies on question-answer pairs, it does not always emphasize learning to extract relevant information from the context provided alongside the question. In standard fine-tuning, the model is trained to directly map a question to its answer without fully leveraging any external context that might be available. As a result, the model may struggle to utilize background information effectively, especially when answering questions that require specialized or up-to-date knowledge.

To address this limitation, we adopt a RAG-style fine-tuning approach using a dataset consisting of \textit{(context, question, answer)} triples. Unlike standard QA fine-tuning, which relies solely on the question, our method introduces an external relevant passage \( p \) for the input question \( q \). This enables the model to learn how to incorporate relevant contextual information during answer generation.
Both approaches share the same next-token prediction objective, but differ in the input they condition on. In standard fine-tuning, the model is trained as follows:
\[
\mathcal{L}_{\text{QA Fine-tuning}} = - \sum_{i=1}^{n} \log p_\theta(x_i \mid [q; x_{<i}]),
\]
where the model predicts each answer token \( x_i \) based only on the question and the previously generated tokens.
In contrast, RAG-style fine-tuning conditions the generation not only on the question but also on the relevant passages as context:
\[
\mathcal{L}_{\text{RAG Fine-tuning}} = - \sum_{i=1}^{n} \log p_\theta(x_i \mid [q, c; x_{<i}]),
\]
where \( c \) is the relevant passage retrieved from an external corpus. By incorporating \( c \) as an additional context, the model is encouraged to utilize external knowledge when generating answers. This strategy improves the model’s ability to ground its responses in retrieved evidence, leading to more accurate and contextually appropriate answers. During RAG fine-tuning, we used gold passages with high relevance to the answers, ensuring the model learns to effectively utilize contextual information.

\section{Dataset}

\subsection{MusWikiDB}

To address the lack of a music-specific vector database for RAG in MQA, we developed \textbf{MusWikiDB}. We began by collecting music-related content from Wikipedia across seven categories: \textit{artists, genres, instruments, history, technology, theory}, and \textit{forms}. These categories were selected to cover a broad spectrum of music knowledge, providing a well-rounded foundation for answering music-related questions. The data was collected with a page depth of 2, which allowed us to capture detailed subtopics and related information. We split the content into sections such as \textit{background, biography}, and \textit{history}. We then removed sections shorter than 60 tokens to ensure the remaining text had enough context for meaningful retrieval. 

Table~\ref{tab:db_statistics} compares our proposed MusWikiDB with the Wikipedia corpus~\cite{karpukhin2020dense}. While MusWikiDB contains fewer pages (31K vs 3.2M) and has a smaller vocabulary size (786K vs 21.5M), it consists exclusively of music-specialized text information.

% 아래 내용 추가해야함.
% 1. 마지막 청크의 경우 60토큰 미만이어도 포함했음

Based on the ablation study in Section ~\ref{sec:ablation}, the text was then split into segments of up to 128 tokens, with a 10\% overlap between adjacent passages, to preserve context between passages. For embedding, we employed BM25~\cite{robertson1994some}, a classical and highly effective algorithm for ranking text relevance, which helped build an efficient index for MusWikiDB. This allowed us to quickly retrieve relevant information during RAG-based inference, improving the accuracy and relevance of answers. The resulting MusWikiDB provides a scalable, up-to-date knowledge base that enhances the performance of RAG in MQA tasks, allowing the system to answer complex, domain-specific music-related questions with more accuracy and context. 
% Detailed statistics of MusWikiDB can be found in Table~\ref{tab:db_statistics}.

\begin{table}[!t]
\centering
% \resizebox{\linewidth}{!}{
\begin{tabular}{@{}lcc@{}}
\toprule
\multicolumn{1}{l}{} & \textbf{MusWikiDB} & \textbf{Wikipedia Corpus}~\cite{karpukhin2020dense} \\ \midrule
\# Pages             & 31K       & 3.2M             \\
\# Passages          & 629.2K    & 21M              \\
Total tokens         & 65.5M     & 2.1B             \\
Vocab Size           & 786K      & 21.5M            \\ \bottomrule
\end{tabular}
% }
\caption{MusWikiDB and Wikipedia Corpus~\cite{karpukhin2020dense} statistics.}
\vspace{-6mm}
\label{tab:db_statistics}
\end{table}

% \begin{table}[h]
% \centering
% \resizebox{0.4\linewidth}{!}{
% % \begin{tabular}{@{}cc@{}}
% % \toprule
% % \textbf{Metric} & \textbf{Value} \\ \midrule
% % \# Data         & 629K        \\
% % Total tokens    & 65.5M          \\
% % Vocab Size      & 786K           \\
% % Avg tokens      & 104            \\
% % Max tokens      & 128            \\
% % Min tokens      & 60            \\ \bottomrule
% % \end{tabular}
% % }
% \caption{\textbf{MusWikiDB} statistics.}
% \label{tab:muswikidb_statistics}
% \end{table}

\subsection{ArtistMus}

% The existing text-only MQA benchmarks have focused on music understanding and generation, as well as musicology topics such as melody, chords, and history~\cite{yuan2024chatmusician, ramoneda2024role}. 
The existing text-only MQA benchmarks have focused on multimodal music understanding~\cite{weck2024muchomusic, yuan2024chatmusician} or musicology topics such as melody, chords, and history~\cite{yuan2024chatmusician, ramoneda2024role}. However, there has been no benchmark that focuses on music metadata, particularly the artist, which is crucial in music listening contexts~\cite{lee2010analysis, doh2024music}. Therefore, we created the \textbf{ArtistMus} to test the performance of LLMs in artist-related QA, using artist-related data from MusWikiDB.

%We crawled music artists from Wikipedia who have genre and year information in their infoboxes. 
We grouped sections into five categories: \textit{biography, career, discography, artistry, and collaborations}. Token lengths ranging from 500 to 2000 were considered. Genre normalization~\cite{schreiber2015improving} was applied by first converting all genre labels to lowercase, and then removing spaces, hyphens (-), and slashes (/).
We obtained 48 root genres from ~\cite{schreiber2016genre}, and after retaining only the data corresponding to the top 300 most frequent genres, each genre was mapped to the 20 final genre labels. 
%These 20 genres align well with the 16 genres in \cite{schreiber2015improving}.
To extract artists' regional information, we provided the abstract of pages to the Llama 3.1 8B Instruct~\cite{grattafiori2024llama} to extract information on the country of the artist. The country list was obtained from the \textit{pycountry} library. 
%for cases like \textit{Dutch $\rightarrow$ Netherlands} and \textit{UK $\rightarrow$ United Kingdom}. %The prompt for country extraction is presented in Table \ref{tab:country_extract_prompt}.
Then, we select a diverse range of 500 artists based on \textit{topic, genre}, and \textit{country}. Country was set as the highest priority, with a preference for artists from minor countries. Subsequently, popular genres and topics were replaced with less common ones.
We generated one factual and one contextual question for each artist to evaluate the LLM's factuality and contextual understanding. To construct these questions, we provided GPT-4o~\cite{achiam2023gpt} with the corresponding section text. Factual questions focus on verifiable details such as dates, names, or events, whereas contextual questions require reasoning or synthesis across multiple pieces of information within the passage.

%Prompts for MCQ generation is presented in Section \ref{sec:mcq_generate}.

We validate the generated questions based on two criteria: \textit{Music Relevance} and \textit{Faithfulness}. For Music Relevance, questions that did not pertain to musical aspects were excluded except important details such as the artist's birthplace. For Faithfulness, GPT-4o was asked to verify whether the question and answer could be derived from the provided text. 
%This process ensured that the generated content remained grounded in the given source material.
Finally, 1,000 multiple-choice questions passing human validation were generated.
%An interesting observation was that the distribution of correct answer choices was highly biased. Specifically, the correct answers were distributed as follows: 'B': 589, 'C': 292, 'A': 113, and 'D': 6 out of 1,000 questions, indicating a significant imbalance. 
We randomly reassigned the correct answers, ensuring an even distribution by assigning 250 correct answers to each option.

\section{Experiments}

\begin{table*}[h]
\centering
\begin{tabular}{@{}lclccclccc@{}}
\toprule
                       & \multicolumn{1}{l}{} &  & \multicolumn{3}{c}{\textbf{Factual}}                &  & \multicolumn{3}{c}{\textbf{Contextual}}             \\ 
\textbf{Model}         & \textbf{Params}      &  & \textbf{Seen}   & \textbf{Unseen} & \textbf{All}    &  & \textbf{Seen}   & \textbf{Unseen} & \textbf{All}    \\ \midrule
\multicolumn{10}{l}{{\color[HTML]{666666} \textit{Baseline Models (zero-shot)}}}                                                                                \\ 
GPT-4o~\cite{achiam2023gpt}                 & N/A                  &  & 70.0          & 64.8          & 67.4          &  & \textbf{93.2} & \textbf{92.8} & \textbf{93.0} \\
ChatMusician~\cite{yuan2024chatmusician}           & 7B                   &  & 28.0          & 25.2          & 26.6          &  & 78.8          & 67.6          & 73.2          \\
MuLLaMA~\cite{liu2024music}                & 7B                   &  & 27.2          & 25.2          & 26.2          &  & 38.4          & 40.0          & 39.2          \\
Llama 3.1 8B Instruct~\cite{grattafiori2024llama}     & 8B                   &  & 40.0          & 38.0          & 39.0          &  & 87.6          & 82.8          & 85.2          \\ \midrule
\multicolumn{10}{l}{{\color[HTML]{666666} \textit{Domain Adaptation Models (Llama 3.1 8B Instruct)}}}                                                           \\
QA Fine-tuning         & 8B                   &  & 41.2          & 38.8          & 40.0          &  & 81.6          & 78.8          & 79.7          \\
RAG Inference (Ours)   & 8B                   &  & 81.2          & 82.8          & 82.0          &  & 89.6          & 88.0          & 88.8          \\
RAG Fine-tuning (Ours) & 8B                   &  & \textbf{81.6} & \textbf{83.2} & \textbf{82.4} &  & 92.4          & 91.6          & 92.0          \\ \bottomrule
\end{tabular}
\caption{Performance on the \textbf{ArtistMus} benchmark. \textit{Seen} refers to data with artists present in training data, while \textit{Unseen} contains new artists. This distinction applies only to domain adaptation models. For baseline models, all data is unseen.}
\label{tab:main_table_zeroshot}
\end{table*}

\begin{table*}[!t]
\centering
\begin{tabular}{@{}lclcccclc@{}}
\toprule
\textbf{Model}       & \textbf{Params} &  & \textbf{Ppl} & \textbf{IT} & \textbf{GFT} & \textbf{CH} &  & \textbf{All}    \\ \midrule
\multicolumn{9}{l}{{\color[HTML]{666666} \textit{Baseline Models (zero-shot)}}}                                                            \\
GPT-4o~\cite{achiam2023gpt}                & N/A             &  & \textbf{48.0}   & \textbf{47.0}       & \textbf{57.0}  & \textbf{60.0}    &  & \textbf{53.0} \\
ChatMusician~\cite{yuan2024chatmusician}         & 7B              &  & 18.0            & 20.0                & 26.0           & 24.0             &  & 20.0          \\
MuLLaMA~\cite{liu2024music}              & 7B              &  & 25.0            & 15.0                & 18.0           & 21.0             &  & 19.8          \\
Llama 3.1 8B Instruct~\cite{grattafiori2024llama}      & 8B              &  & 36.0            & 24.0                & 41.0           & 42.0             &  & 35.8          \\ \midrule
\multicolumn{9}{l}{{\color[HTML]{666666} \textit{Domain Adaptation Models (Llama 3.1 8B Instruct)}}}                                             \\
QA Fine-tuning                  & 8B              &  & 32.0            & 21.0                & 39.0           & 36.0             &  & 32.0          \\
RAG Inference (Ours) & 8B              &  & 33.0            & 40.0                & 44.0           & 46.0             &  & 40.8          \\
RAG Fine-tuning (Ours)        & 8B              &  & 33.0            & 38.0                & 46.0           & 49.0             &  & 41.5          \\ \bottomrule
\end{tabular}
\caption{Performance on out-of-domain (OOD) \textbf{TrustMus} benchmark. Four categories are: People (Ppl), Instrument \& Technology (IT), Genre, Forms, and Theory (GFT), and Culture \& History (CH).}
\label{tab:trustmus_result}
\end{table*}

\subsection{Benchmarks}

For evaluation, we used two datasets: ArtistMus (in-domain) and TrustMus (out-of-domain). Performance on factual and contextual questions was separately measured on the ArtistMus. For TrustMus, evaluation was conducted across four categories: People (Ppl), Instrument \& Technology (IT), Genre, Forms, and Theory (GFT), and Culture \& History (CH), each comprising 100 questions. All evaluations use a multiple-choice QA format.

\subsection{Models}
We compare zero-shot and QA fine-tuned models with our proposed RAG inference and RAG fine-tuned models to evaluate MQA performance. Following \cite{weck2024muchomusic}, we consider a response incorrect if it deviates from the expected format.

\vspace{1mm} \noindent \textbf{Zero-shot Baselines}~~
We evaluated GPT-4o~\cite{achiam2023gpt} (API-based), Llama 3.1 8B Instruct~\cite{grattafiori2024llama} (open-source), and two music-specific models: MuLLaMA~\cite{liu2024music} and ChatMusician~\cite{yuan2024chatmusician}. MuLLaMA is designed to handle audio based question answering. ChatMusician specializes in music understanding and generation with ABC notation.

% MuLLaMA is designed to handle \sh{audio understanding questions}. ChatMusician focuses on modeling music as a language to understand and generate music content through text-based interactions. These models were evaluated to examine how music-specialized LLMs perform on MQA tasks.

\vspace{1mm} \noindent\textbf{QA Fine-tuning}~~
We fine-tune the Llama 3.1 8B Instruct~\cite{grattafiori2024llama} on 8K multiple-choice QA pairs that were generated from MusWikiDB.
% by the same model itself.

\vspace{1mm} \noindent\textbf{RAG Inference}~~
We use Llama 3.1 8B Instruct~\cite{grattafiori2024llama} as our base model and implement RAG at inference-time using MusWikiDB as the retrieval database.

\vspace{1mm} \noindent\textbf{RAG Fine-tuning}~~
We performed RAG fine-tuning using a dataset in the form of \textit{(context, question, answer)}, by augmenting the original QA fine-tuning dataset with additional context. The target model and all other training settings were kept identical to those used in QA fine-tuning.

\subsection{Training Configurations}
The models are trained for one epoch using LoRA~\cite{hu2022lora} with 8-bit quantization with the following hyperparameter settings: batch size = 2, gradient accumulation steps = 4, learning rate = 3e-5, weight decay = 0.005, warmup ratio = 0.1, cosine scheduler~\cite{wolf-etal-2020-transformers}, AdamW~\cite{loshchilov2017decoupled} optimizer, r = 16, alpha = 16, and dropout = 0.1. For the ArtistMus dataset, half of the artists were included in the training data (Seen), while the other half were excluded (Unseen).

\subsection{Retriever Configurations}
To select the optimal retriever configuration MusWikiDB, we performed an ablation study using the ArtistMus benchmark. We varied the passage size (128, 256, 512 tokens) and embedding models (BM25~\cite{robertson1994some}, Contriever~\cite{izacard2021unsupervised}, CLAP~\cite{wu2023large}). For CLAP, we increased the token limit without additional training.
To ensure a fair comparison, we constrained the total token budget to 1024 by adjusting the number of retrieved passages: top-8 for 128-token passages, top-4 for 256, and top-2 for 512. 

% The best configuration was selected based on overall performance. We also report the gold performance as an upper bound, where the source text used to create each benchmark QA is provided.

\section{Result}

\subsection{In-domain Performance}

\noindent \textbf{Zero-shot Baselines}~~
As shown in Table~\ref{tab:main_table_zeroshot}, all models performed significantly worse on factual questions than on contextual ones, indicating challenges in recalling concrete information such as names or dates. GPT-4o~\cite{achiam2023gpt} outperformed Llama~\cite{grattafiori2024llama} by 28.4\% in factual performance, though the gap narrowed to 7.8\% for contextual understanding. Despite being music-specific, both ChatMusician~\cite{yuan2024chatmusician} and MuLLaMA~\cite{liu2024music} showed relatively low performance. ChatMusician slightly underperformed compared to Llama, while MuLLaMA exhibited the lowest scores, likely due to its lack of training on the MQA task and poor instruction-following capabilities.

\vspace{1mm} \noindent \textbf{QA Fine-tuning}~~
%The bottom rows of Table \ref{tab:main_table_zeroshot} presents the results of QA fine-tuned and RAG fine-tuned models. 
Comparing QA fine-tuning with zero-shot performance, factual performance improved by 1.0\%, but contextual performance decreased by 5.5\%. This suggests that while QA fine-tuning is effective in helping the model retain information from the training data, it may also reduce the overall inference capability.

\vspace{1mm} \noindent \textbf{RAG Inference}~~
By utilizing RAG inference without additional training, we were able to address the low factual performance that was an issue with previous LLMs. It demonstrated a 14.6\% higher factual performance compared to GPT-4o~\cite{achiam2023gpt}. Contextual performance improved by 3.6\% compared to zero-shot, but was still 4.2\% lower than GPT-4o.

\vspace{1mm} \noindent \textbf{RAG Fine-tuning}~~
The model fine-tuned on the RAG-style dataset showed improvements in both types of questions. Compared to RAG inference, factual performance improved by 0.4\%, and contextual performance improved by 3.2\%. This demonstrates that by learning to leverage context, the model not only improves its memory of information present in the training data but also enhances its overall contextual understanding ability. It exhibited a remarkable 15.0\% higher factual performance compared to GPT-4o, and only 1.0\% lower contextual performance, which is nearly equivalent. Considering factors such as the model size, amount of training data, and the extent of training, this is an exceptionally high performance.

\subsection{Out-of-domain Performance}
To validate the effectiveness of the MusT-RAG in out-of-domain scenarios, we conducted experiments by changing the benchmark from ArtistMus to TrustMus~\cite{ramoneda2024role}, using the same framework with in-domain evaluation. The results are presented in Table ~\ref{tab:trustmus_result}.

\vspace{1mm} \noindent \textbf{Zero-shot Baselines}~~
A similar trend was observed in the zero-shot evaluation for the in-domain setting. MuLLaMA~\cite{liu2024music} and ChatMusician~\cite{yuan2024chatmusician} performed worse than the random baseline (25\%), which is due to incorrect answers being counted when the models failed to follow instructions. Given that the overall zero-shot performance closely aligns with the factual scores from the in-domain evaluation, we infer that TrustMus mostly consists of factual questions. The Llama 3.1 8B Instruct~\cite{grattafiori2024llama} model scored 17.2\% lower than GPT-4o~\cite{achiam2023gpt}.

\vspace{1mm} \noindent \textbf{QA Fine-tuning}~~
The QA fine-tuned model showed a 3.8\% decrease in performance compared to zero-shot, which can be attributed to the fact that models trained on artist data tend to forget information about out-of-domain topics, such as \textit{Instrument} and \textit{Genre}.

\vspace{1mm} \noindent \textbf{RAG Inference}~~
RAG inference led to an 5.0\% performance improvement over zero-shot, demonstrating that MusT-RAG framework is also helpful for out-of-domain data, such as \textit{The Grove Dictionary Online}~\cite{sadie2001new}, which is the basis for the TrustMus benchmark.

\vspace{1mm} \noindent \textbf{RAG Fine-tuning}~~
The RAG fine-tuned model showed a 0.7\% improvement over RAG inference, even with the same artist data used for QA fine-tuning. This supports the fact that the RAG fine-tuning method, which incorporates context, enhances the model's robustness in contextual understanding, even for out-of-domain data.

% To validate the effectiveness of the MusT-RAG in out-of-domain scenarios, we conducted experiments by changing the benchmark from ArtistMus to TrustMus~\cite{ramoneda2024role}, using the same framework with in-domain evaluation. The results are presented in Table ~\ref{tab:trustmus_result}.
% The QA fine-tuned model showed a 3.8\% decrease in performance compared to zero-shot, which can be attributed to the fact that models trained on artist data tend to forget information about out-of-domain topics, such as \textit{Instrument} and \textit{Genre}. On the other hand, RAG inference led to an 5.0\% performance improvement over zero-shot, demonstrating that MusT-RAG framework is also helpful for out-of-domain data, such as \textit{The Grove Dictionary Online}~\cite{sadie2001new}, which is the basis for the TrustMus benchmark. The RAG fine-tuned model showed a 0.7\% improvement over RAG inference, despite the training data being the same artist data as in the QA fine-tuned model. This supports the fact that the RAG fine-tuning method, which incorporates context, enhances the model's robustness in contextual understanding, even for out-of-domain data.

\begin{table}[!t]
\centering
\resizebox{\linewidth}{!}{
\begin{tabular}{@{}lccc@{}}
\toprule
\textbf{Embedding}          & \textbf{Passage Size} & \textbf{Factual} & \textbf{Contextual} \\ \midrule 
\multicolumn{2}{l}{Gold (Upper Bound)}                  & 97.8           & 97.0      \\ \midrule
\multirow{3}{*}{BM25~\cite{robertson1994some}}       & 512                 & 82.0          &88.8      \\
                            & 256                 & \textbf{82.8}           & 88.0                \\
                            & 128                 & 82.2           & \textbf{89.0}                \\ \midrule
\multirow{3}{*}{Contriever~\cite{izacard2021unsupervised}} & 512                 & 46.6           & 81.0                \\
                            & 256                 & 55.6           & 84.2                \\
                            & 128                 & 58.2  & 86.6                \\ \midrule
\multirow{3}{*}{CLAP~\cite{wu2023large}}       & 512                 & 41.2  & 79.6                \\
                            & 256                 & 41.0           & 84.0                \\
                            & 128                 & 41.8           & 84.0                \\ \bottomrule
\end{tabular}
}
\caption{Llama 3.1 8B Instruct~\cite{grattafiori2024llama} RAG performance on \textbf{ArtistMus}, by different passage size and embeddings.}
\label{tab:rag_chunk_embedding}
\vspace{-1.5mm}
\end{table}

\subsection{Ablation Study: Retriever Configurations}~\label{sec:ablation}
\begin{figure}[!t]
  \centering
  \captionsetup{skip=1.5pt}
  \includegraphics[width=\columnwidth]{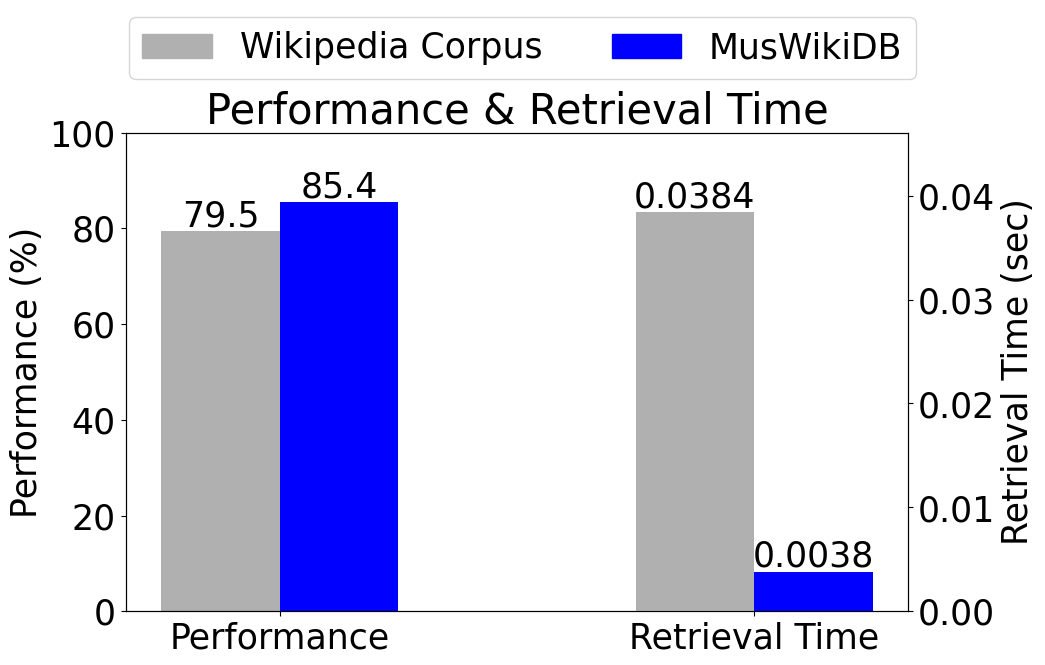}
  \caption{RAG performance and retrieval time for Wikipedia Corpus~\cite{karpukhin2020dense} and MusWikiDB.}
  \label{fig:db_difference}
  \vspace{-1em}
\end{figure}

Table~\ref{tab:rag_chunk_embedding} shows the results of the RAG inference for the Llama 3.1 8B Instruct~\cite{grattafiori2024llama} with various passage sizes and embeddings, evaluated under the same total computation budget for fair comparison. The performance on contextual questions tended to improve as the passage size decreased across all embedding models. In contrast, for factual questions, only Contriever~\cite{izacard2021unsupervised} showed clear improvements with shorter passages, while BM25~\cite{robertson1994some} and CLAP~\cite{wu2023large} showed little to no change in performance across different passage lengths. For factual questions, there was a significant performance gap between BM25 and the other two dense embeddings. This is likely because ArtistMus places high importance on music entities such as artist and albums. Overall, the best performance was achieved using BM25 with a passage size of 128. When compared to the gold context, the factual performance was 15.6\% lower, and the contextual performance was 8.0\% lower. In Figure~\ref{fig:db_difference}, we compare the RAG inference performance using the Wikipedia corpus~\cite{karpukhin2020dense} and MusWikiDB. The results show that MusWikiDB achieves a 10x faster retrieval speed and 5.9\% higher performance.

\section{Conclusion}

% In this paper, we propose the \textbf{MusT-RAG} framework, which enhances text-only Music Question Answering (MQA) by leveraging Retrieval-Augmented Generation (RAG) as a domain adaptation technique for general Large Language Models (LLMs). %A key component of this framework is MusWikiDB, the first vector database specifically designed to cover a wide range of music-related topics. 

% % By integrating MusWikiDB into the MusT-RAG approach, we address the critical issue of low factual recall in existing LLMs. 
% Our experiments show that the RAG inference of Llama 3.1 8B Instruct~\cite{grattafiori2024llama} model hugely outperforms GPT-4o~\cite{achiam2023gpt} for factual recall, demonstrating the effectiveness of our approach. Furthermore, to tackle the challenge of reduced contextual understanding often seen in traditional QA fine-tuning, we introduce a RAG-style fine-tuning process that trains the model to better handle context, leading to improvements in both factual recall and contextual understanding. Our RAG fine-tuned model outperformed GPT-4o by 22.6\% in factual recall, while demonstrating nearly identical performance in contextual understanding. Finally, our framework has proven its versatility by achieving strong results on the out-of-domain benchmark, TrusMus~\cite{ramoneda2024role}. Our MusWikiDB and ArtistMus datasets will serve as valuable resources for the future application of the MusT-RAG approach when the domain adaptation of LLMs in the music domain is needed.

%%% 아래는 GPT 버전 %%%

In this paper, we presented \textbf{MusT-RAG}, a retrieval-augmented framework that enhances text-only Music Question Answering (MQA) by adapting general-purpose LLMs to the music domain. By retrieving relevant passages from a music-specific database and incorporating them into the generation context, MusT-RAG effectively mitigates the factuality limitations commonly observed in LLMs. As a result, our method achieves substantial improvements over GPT-4o~\cite{achiam2023gpt}, particularly in factual accuracy.
Beyond simple retrieval, we further demonstrated that RAG-style fine-tuning outperforms traditional QA fine-tuning by improving both factual and contextual performance. Our final model achieves a 15.0\% gain in factual performance over GPT-4o while maintaining comparable performance in contextual tasks.
Importantly, MusT-RAG shows strong generalization capabilities. On the out-of-domain benchmark TrustMus~\cite{ramoneda2024role}, it delivers a 5.7\% performance improvement over the zero-shot baseline, underscoring its robustness across diverse music-related QA scenarios.
To facilitate future work in this underexplored domain, we release two key resources: \textbf{MusWikiDB}, a music-specific retrieval corpus, and \textbf{ArtistMus}, a benchmark focused on artist-level musical knowledge. We hope these contributions will drive further progress in developing accurate and domain-aware LLMs for music understanding and beyond.

\newpage

\newpage

\newpage

\bibliography{ISMIRtemplate}

% For non BibTeX users:
%\begin{thebibliography}{citations}
% \bibitem{Author:17}
% E.~Author and B.~Authour, ``The title of the conference paper,'' in {\em Proc.
% of the Int. Society for Music Information Retrieval Conf.}, (Suzhou, China),
% pp.~111--117, 2017.
%
% \bibitem{Someone:10}
% A.~Someone, B.~Someone, and C.~Someone, ``The title of the journal paper,''
%  {\em Journal of New Music Research}, vol.~A, pp.~111--222, September 2010.
%
% \bibitem{Person:20}
% O.~Person, {\em Title of the Book}.
% \newblock Montr\'{e}al, Canada: McGill-Queen's University Press, 2021.
%
% \bibitem{Person:09}
% F.~Person and S.~Person, ``Title of a chapter this book,'' in {\em A Book
% Containing Delightful Chapters} (A.~G. Editor, ed.), pp.~58--102, Tokyo,
% Japan: The Publisher, 2009.
%
%\end{thebibliography}

\end{document}